\documentclass[11pt]{article}

\usepackage[preprint]{acl}

\usepackage{times}
\usepackage{latexsym}

\usepackage[T1]{fontenc}

\usepackage[utf8]{inputenc}

\usepackage{microtype}

\usepackage{inconsolata}

\usepackage{graphicx}

\usepackage{amsmath}
\usepackage{amssymb}
\usepackage{amsthm}
\usepackage{booktabs}
\usepackage{multirow}
\usepackage{xcolor}
\usepackage{colortbl}
\usepackage{subcaption}
\usepackage{algorithm}
\usepackage{algorithmic}
\usepackage{tikz}
\usetikzlibrary{matrix,positioning,fit,backgrounds,calc}
\usepackage{enumitem}
\usepackage{url}

\definecolor{lightgray}{gray}{0.92}
\definecolor{bestcolor}{HTML}{E8F5E9}

\newcommand{\ours}{\textsc{HRBench}}
\newcommand{\think}{\texttt{think}}
\newcommand{\nothink}{\texttt{no\_think}}
\newcommand{\PT}{\textsc{PT}}
\newcommand{\RT}{\textsc{RT}}  
\newcommand{\Spec}{\textsc{Spec}}

\newtheorem{definition}{Definition}
\newtheorem{strategy}{Strategy}
\newtheorem{problem}{Problem}

\title{HRBench: Benchmarking and Understanding Thinking-Mode Switch \\ Strategies in Hybrid-Reasoning LLMs}

\author{
Yansong Ning$^{1}$ \thanks{Work done during an internship at Tencent.} \quad
Mianpeng Liu$^{1}$ \quad
Jingwen Ye$^{2}$ \quad
Weidong Zhang$^{2}$ \quad
Hao Liu$^{1}$ \thanks{Corresponding author.}  \\
$^{1}$ AI Thrust, The Hong Kong University of Science and Technology (Guangzhou) \\
$^{2}$ AIPD, Tencent \\
\texttt{\{yning092,mliu603,liuh\}@hkust-gz.edu.cn} \\
\texttt{\{jingwenye,wadewdzhang\}@tencent.com}
}

\begin{document}
\maketitle

\begin{abstract}
Hybrid-reasoning large language models (LLMs) expose explicit controls over reasoning effort, allowing users or systems to trade off answer quality against inference cost. However, existing methods for adaptive thinking-mode selection are typically evaluated under different models, datasets, and implementation assumptions, making it difficult to compare their practical behavior. We introduce HRBench, a unified evaluation framework for studying thinking-mode switching in hybrid-reasoning LLMs. HRBench organizes the design space along two axes: three switching strategy families, prompt-based selection, external routing, and speculative execution, and four training regimes, training-free, SFT, offline and online RL, yielding 12 controlled evaluation settings. We evaluate these settings across 6 LLMs, from Qwen3.5-2B to Kimi-K2.5-1.1T, and 5 reasoning benchmarks covering mathematics, science, and code, while reimplementing 12+ representative prior methods within the same pipeline. Our analysis characterizes how different switching strategies occupy distinct effectiveness-efficiency trade-off regions: prompt-based methods often provide favorable token-accuracy trade-offs, routing methods offer more stable cost reduction, and speculative methods tend to improve accuracy at higher token cost. We further find that training affects strategies differently, and that the preferred strategy varies with model scale and task domain. HRBench provides reference implementations and a unified evaluation platform to support more controlled research on efficient reasoning in hybrid-reasoning LLMs.
Our data, code and repository are available at \href{https://github.com/usail-hkust/HRBench}{https://github.com/usail-hkust/HRBench}.

\end{abstract}

\section{Introduction}
\label{sec:intro}

Recent reasoning-oriented LLMs, such as OpenAI o1 \citep{openai2024o1} and DeepSeek-R1 \citep{deepseek2025r1}, are achieving remarkable success on complex tasks through extended chain-of-thought (CoT) reasoning \cite{wei2022chain}, but at the cost of substantial token overhead. To address this, a new generation of \emph{hybrid-reasoning LLMs} has emerged, including Qwen3.5 \citep{qwen2025qwen3}, gpt-oss \cite{agarwal2025gpt}, and Seed-OSS \cite{bytedance2025seedoss}, that expose explicit \emph{thinking-mode switches}: users can select between deep reasoning (\think{}) and direct answering (\nothink{}), specify discrete reasoning effort levels (e.g., High/Medium/Low), or set numeric budgets (e.g., $\leq 4096$ tokens). This raises a question: \textbf{when should the model think, and how much?}

\begin{figure*}[t]
    \centering
    \includegraphics[width=\textwidth]{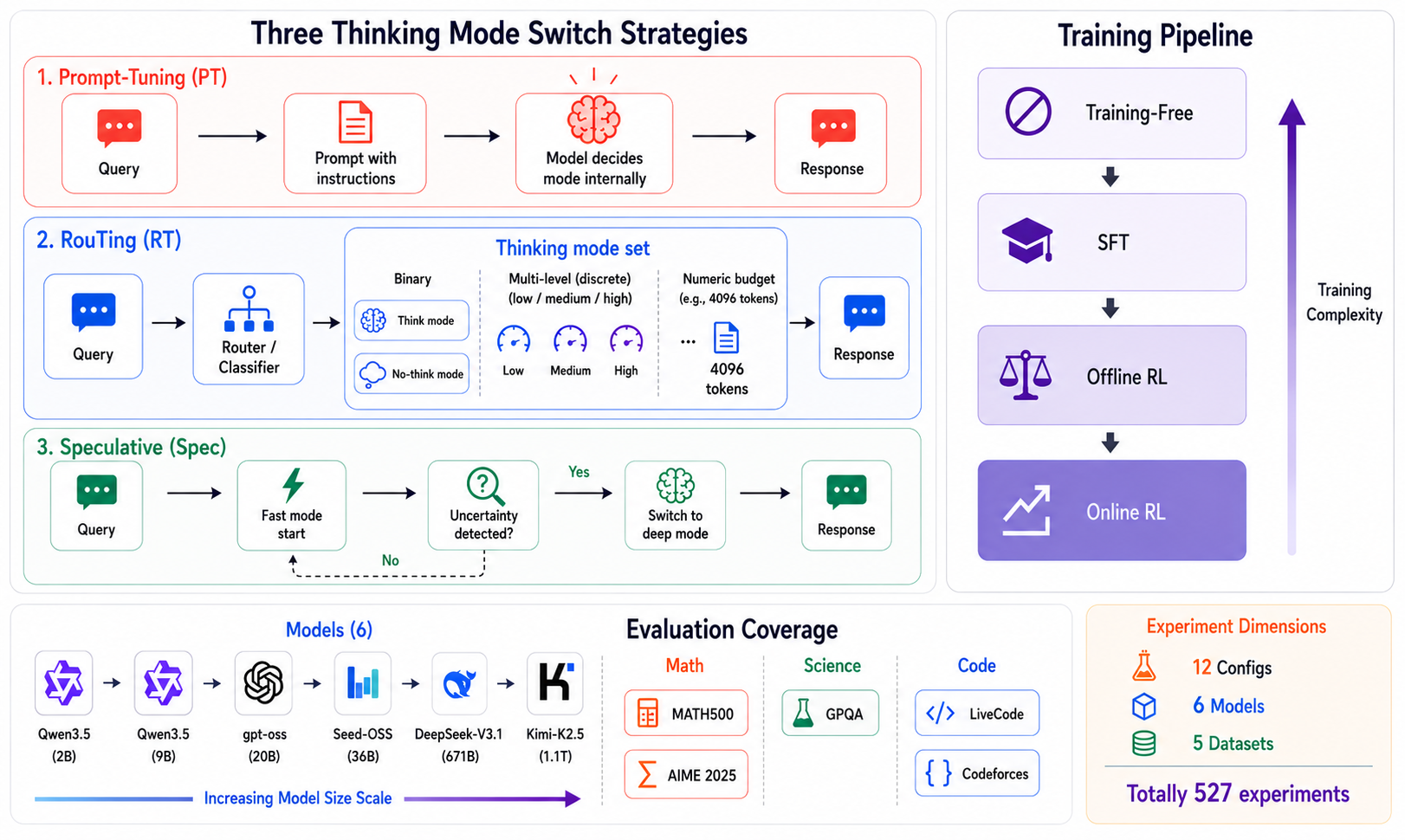}
    \caption{Overview of \ours{}. \textbf{Left}: Three thinking-mode switch strategies---Prompt-Tuning, Routing, and Speculative. \textbf{Right}: Training pipeline spanning training-free to online RL. \textbf{Bottom}: Our evaluation coverage across 6 models spanning from 2B to 1.1T scale, 5 datasets, totaling 527 experiment runs.}
    \label{fig:overview}
\end{figure*}
A growing body of work tackles this efficiency--effectiveness trade-off by proposing \emph{adaptive thinking-mode switch} methods. These can be categorized into three thinking-mode switch strategies:
\begin{itemize}[nosep,leftmargin=*]
    \item \textbf{Prompt-Tuning (\PT{})} guides the model to determine its thinking mode during a single inference pass through carefully designed prompts. For example, methods such as S1 \citep{muennighoff2025s1} and TALE \citep{han2025tale} inject token-budget or difficulty-aware instructions that let the model control reasoning length. Furthermore, RL-based approaches like ACPO \citep{cheng2025acpo} directly optimize this internal decision.
    \item \textbf{RouTing (\RT{})} adopts a classify-then-generate strategy, where a router evaluates the query difficulty before dispatching it to the appropriate thinking mode.
    As representative examples, AdaptThink \citep{zhang2025adaptthink} trains such a router via GRPO, while HDFlow \citep{yao2024hdflow} uses rule-based difficulty classification.
    \item \textbf{Speculative (\Spec{})} methods allow the model to begin in a fast mode and dynamically switch to deep reasoning upon detecting uncertainty signals. For instance, MixReasoning \citep{lu2025mixreasoning} uses entropy-based triggers for this escalation, while ADR \citep{zhang2025adr} learns the switching policy through SFT and RL.
\end{itemize}
Despite active progress, these methods are evaluated under incomparable settings---different LLMs, datasets, metrics, and decoding configurations---making it impossible to answer: \emph{which strategy truly works best?} and \emph{how much does the training process help each strategy?}

To address this, we propose \ours{} shown in Figure~\ref{fig:overview}, a unified benchmark for understanding how different thinking-mode switch strategies behave across strategies, training regimes, model scales, and task domains. We orthogonally combine the three strategies with both training-free and training-based (SFT, offline RL, online RL) approaches, yielding 12 evaluation configurations that cover representative existing methods. 
Under a unified pipeline---6 LLMs spanning Qwen3.5-2B to Kimi-K2.5-1.1T, 5 benchmarks covering math, code, and science, and unified metrics---we systematically characterize how strategies navigate the efficiency--effectiveness trade-off, how training signals interact with strategy choice, and how the optimal strategy--training combination shifts with model scale and task domain. 
Then, we further integrate 12+ existing thinking mode switch methods into the pipeline, providing the unified comparison across the full taxonomy.

Overall, our contributions are summarized in three aspects as follows:
\begin{itemize}[nosep,leftmargin=*]
    \item \textbf{Unified evaluation framework.} We present the first benchmark that systematically covers 12 configurations for the thinking-mode switch, enabling controlled cross-strategy comparison under identical conditions.

    \item \textbf{Systematic empirical analysis of  thinking switch mechanisms.} We reveal that:
    \begin{itemize}[nosep]
        \item The three strategy exhibit fundamentally different trade-off profiles: \PT{} achieves a ``win-win'' (higher accuracy and fewer tokens), \RT{} offers moderate token savings with preserved accuracy, while \Spec{} improves accuracy at additional token cost.
        \item Training gains are strategy-dependent: GRPO achieves the highest token reduction for \RT{} while all training methods maintain comparable accuracy across strategies.
        \item Both effects shift with LLM size scale and task domain: \Spec{} surpasses \PT{} at the 20B and 671B scales, while \PT{} performs better on math and \Spec{} on code tasks.
    \end{itemize}

    \item \textbf{Open-source baselines and platform.} Reference implementations for all 12 configurations and 12+ integrated prior methods, forming a plug-and-play platform for the community.
\end{itemize}

\section{Related Work}
\label{sec:related}

\subsection{Hybrid-Reasoning LLMs}
\label{sec:related_hybrid}

Recent work has introduced LLMs with user-controllable thinking-mode switches that enable flexible allocation of inference compute \citep{wang2025demystifying}. These \emph{hybrid-reasoning LLMs} offer three forms of control over reasoning depth:
\begin{itemize}[nosep,leftmargin=*]
\item \textbf{Binary switch.} The most common way provides a \think{}/\nothink{} switch, where the former activates extended chain-of-thought and the latter generates direct answers. 
Current LLMs adopting this design include Qwen3.5 \citep{qwen2025qwen3}, DeepSeek-V3.1, Kimi-K2.5, and so on.

\item \textbf{Discrete reasoning effort.}
Certain LLMs expose tiers of reasoning effort, e.g, gpt-oss-20B \citep{agarwal2025gpt} introduces High/Medium/Low settings. 
This approach affords coarse-grained control of test-time compute.

\item \textbf{Numeric budget.}
LLM family like Seed-OSS-36B \citep{bytedance2025seedoss} also accept explicit token budgets $b \leq B_{\max}$, enabling fine-grained, continuous control over the token of reasoning.
\end{itemize}

\subsection{Adaptive Thinking-Mode Switch}
\label{sec:related_switch}

Existing adaptive thinking-mode switch methods can be categorized into three categories based on when and how the mode decision is made.

\paragraph{Prompt-Tuning.} 
PT-based methods guide mode selection through prompt engineering within a single inference pass---the model itself decides whether and how deeply to reason. Both training-free and training-based approaches have been explored. Training-free approaches include S1 budget forcing \citep{muennighoff2025s1} and TALE token-budget-aware reasoning \citep{han2025tale}. SFT-based methods include OThink-R1 \citep{zhang2025othink} and HGPO \citep{jiang2025thinkonly}. RL-based methods include ACPO \citep{cheng2025acpo}, and Think-Only (HGPO) \citep{jiang2025thinkonly}. DPO-based methods include AdaR1 \citep{luo2025adar1} and Think-in-Blocks \citep{zhu2025thinkinblocks}.

\paragraph{Routing.} 
RT-based methods employ an explicit two-stage process: a router first assesses query difficulty and selects the appropriate mode, then the model generates under that mode. Training-free routers include HDFlow \citep{yao2024hdflow} and CP-Router \citep{su2025cprouter}. SFT-trained routers include Self-Route \citep{he2025self}, and ThinkSwitcher \citep{thinkswitcher2025}. For example, AdaptThink \citep{zhang2025adaptthink} trains a routing policy via GRPO that decides \think{}/\nothink{} per query, while Self-Route uses a lightweight SFT-trained linear classifier on hidden-state features.

\paragraph{Speculative.} 
Spec-based methods dynamically switch modes during inference. 
The model begins in a fast mode (typically \nothink{}) and triggers a switch to deep reasoning upon detecting uncertainty signals mid-stream.
For example, training-free approaches include MixReasoning \citep{lu2025mixreasoning}, which uses entropy-based triggers to detect when the fast-mode output is unreliable. 
In addition, ADR \citep{zhang2025adr} combines SFT and GRPO stages for learned switching policies.

However, these approaches are evaluated in isolation. 
No prior work provides a unified framework that enables systematic cross-strategy comparison under controlled conditions, which is precisely the gap \ours{} addresses.

\section{Preliminary}
\label{sec:preliminary}

\begin{definition}[Thinking Mode]
In a hybrid-reasoning LLM $\pi_\theta$, a thinking mode $m$ is defined as a control parameter that dictates the token budget allocated for intermediate chain-of-thought $\tau$ before generating a final answer $a$. 
\end{definition}

\noindent The set of all available thinking modes, denoted as $\mathcal{M}$, typically takes one of the following forms depending on the model architecture:
\begin{itemize}[nosep,leftmargin=*]
\item A binary state space: $\mathcal{M} = \{\mathrm{think}, \mathrm{no\_think}\}$. 
\item A discrete effort space: $\mathcal{M} = \{\mathrm{low}, \mathrm{mid}, \mathrm{high}\}$.
\item A continuous token budget space: $\mathcal{M} = \{b \mid b \in [0, B_{\max}]\}$, where $b$ specifies the maximum number of intermediate reasoning tokens.
\end{itemize}

\begin{problem}[Thinking Mode Switch]
Given a query $q$ and the thinking modes set $\mathcal{M}$, the model $\pi_\theta$ adaptively selects approximate thinking modes to generate a response $r = (\tau, a)$, where $\tau$ denotes the chain-of-thought and $a$ is the final answer.
\end{problem}

\noindent In this paper, the above problem can be solved by the following three strategies:

\begin{table*}[t]
\centering
\small
\begin{tabular}{l|cccc}
\toprule
& \textbf{Training-Free} & \textbf{SFT} & \textbf{Online RL} & \textbf{Offline RL} \\
\midrule
\textbf{Prompt-Tuning} & S1, TALE & OThink-R1, HFT & AdaR1 & ACPO, HGPO, SABER \\
\textbf{Routing} & HDFlow, CP-Router & Self-Route, ThinkSwitcher & \textcolor{gray}{\emph{Ours}} & AdaptThink \\
\textbf{Speculative} & MixReasoning & ADR$_{\text{SFT}}$ & \textcolor{gray}{\emph{Ours}} & ADR$_{\text{GRPO}}$ \\
\bottomrule
\end{tabular}
\caption{The evaluation taxonomy of the \ours{}. Each cell represents a unique (strategy, training regime) configuration. Representative external methods are listed for each cell. \textcolor{gray}{\emph{Ours}} marks 2 configurations without prior baselines, first explored in this work.}
\label{tab:taxonomy}
\end{table*}

\begin{strategy}[Prompt-Tuning based Switch]
The model $\pi_\theta$ implicitly selects a thinking mode $m$ and generates the response:
\begin{equation}
    r \sim \pi_\theta(\cdot \mid q, T_{\emph{PT}}, m)
\end{equation}
\end{strategy}
\noindent where $T_{\text{PT}}$ is a prompt template that encodes mode-selection instructions, and $m$ is implicitly determined by the model $\pi_\theta$ during inference.

\begin{strategy}[Routing based Switch]
A router first explicitly selects a thinking mode, then the model $\pi_\theta$ generates based on the routed thinking mode:
\begin{equation}
   \quad r \sim \pi_\theta(\cdot \mid q, \hat{m}), \hat{m} = \pi_\psi(q)
\end{equation}
\end{strategy}
\noindent where $\pi_\psi$ is the router policy, mapping the query to a specific thinking mode $\hat{m} \in \mathcal{M}$.

\begin{strategy}[Speculative based Switch]
The model $\pi_\theta$ initiates decoding under an initial thinking mode $m_0$ and monitors the partial output.
Upon a trigger signal, it will switch to an alternative thinking mode $m_t$:
\begin{equation}
    r \sim \begin{cases}
    \pi_\theta(\cdot \mid q, m_0) & \text{if } f(\tau_{1:t}) \text{ not triggered} \\
    \pi_\theta(\cdot \mid q, m_t, \tau_{1:t^*}) & \text{if } f(\tau_{1:t^*}) \text{ triggered}
    \end{cases}
\end{equation}
\end{strategy}
\noindent where $m_0, m_t \in \mathcal{M}$ are distinct thinking modes, $\tau_{1:t} \sim \pi_\theta(\cdot \mid q, m_0)$ is the partial chain-of-thought under the initial thinking mode $m_0$, $f$ is a trigger function \cite{yang2025speculative}, and $t^*$ is the token position at which $f(\tau_{1:t^*})$ is satisfied.

\section{HRBench Construction}
\label{sec:setup}

\subsection{Evaluation Taxonomy}
\label{sec:taxonomy}

We organize the evaluation into a systematic taxonomy (Table~\ref{tab:taxonomy}), crossing three strategies with four training regimes to yield 12 configurations.

\subsection{Datasets}
\label{sec:datasets}

As shown in Table~\ref{tab:datasets}, we evaluate on five benchmarks spanning three task domains:

\begin{itemize}[nosep,leftmargin=*]
    \item \textbf{Mathematics}: AIME 2025 (competition-level math problems) and MATH500 (high school math problems) \cite{lightman2023lets}.
    \item \textbf{Science}: GPQA-Diamond (graduate-level questions ranging from physics, chemistry, to biology) \citep{rein2024gpqa}.
    \item \textbf{Code}: Live Code Bench (LCB) (live programming problems with execution-based evaluation) \citep{jain2024livecodebench} and Codeforces (competition-level programming problems).
\end{itemize}

\subsection{Models}
\label{sec:models}
We evaluate 6 hybrid-reasoning LLMs spanning 2B to 1.1T parameters, covering three thinking modes:

\begin{itemize}[nosep,leftmargin=*]
    \item \textbf{Qwen3.5-2B} and \textbf{Qwen3.5-9B} \citep{qwen2025qwen3}: Binary switch (\think{}/\nothink{}).
    \item \textbf{gpt-oss-20B}: Discrete thinking mode switch (e.g., High/Medium/Low reasoning effort).
    \item \textbf{Seed-OSS-36B-Instruct}: Thinking mode switch via numeric token budget ($b \leq B_{\max}$).
    \item \textbf{DeepSeek-V3.1-671B} \cite{deepseek2025v3}: Binary switch (\think{}/\nothink{}).
    \item \textbf{Kimi-K2.5-1.1T} \cite{team2026kimi}: Binary switch (\think{}/\nothink{}).
\end{itemize}


\begin{table}[t]
\centering
\small
\begin{tabular}{l|ccc}
\toprule
\textbf{Dataset} & \textbf{Domain} & \textbf{\#Prob.} & \textbf{Difficulty} \\
\midrule
MATH500 & Math & 500 & High School \\
AIME 2025 & Math & 30 & Competition \\
GPQA-Dia. & Science & 198 & Graduate \\
LCB & Code & 167 & Graduate \\
Codeforces & Code & 366 & Competition \\
\bottomrule
\end{tabular}
\caption{Overall dataset statistics for the five benchmarks used in \ours{}.}
\label{tab:datasets}
\end{table}
\subsection{Metrics}
\label{sec:metrics}

In this paper, we use accuracy and token cost to investigate the effectiveness-efficiency tradeoff:
\begin{itemize}[nosep,leftmargin=*]
    \item \textbf{Acc}: Pass@1 accuracy (\%).
    \item \textbf{Tok}: Average output token cost (including CoT).
\end{itemize}

\subsection{Baselines and Implementations}
\label{sec:baselines}

\paragraph{Fixed baselines.} \textbf{Full-Think} (always \think{}), \textbf{No-Think} (always \nothink{}), and \textbf{Budget-Aware} (High/Medium/Low reasoning effort tiers).

\paragraph{Our implementations.} For each of the 12 taxonomy cells, we provide a reference implementation using verl \citep{sheng2024verl} for training and vLLM \citep{kwon2023vllm} for inference. All methods are evaluated under identical decoding parameters. 
Details are provided in Appendix~\ref{app:impl_details}. We categorize implementations into two parts:

\begin{itemize}[leftmargin=*]
\item \textbf{Training-Free (TF) Implementation}:
    \begin{itemize}[nosep,leftmargin=*]
    \item \textbf{Prompt-Tuning (PT-TF)}: We craft model-specific prompts mapping to reasoning effort levels (e.g., \texttt{think}/\texttt{no\_think}, token budgets), enabling the LLM to auto-select its mode.
    \item \textbf{Routing (RT-TF)}: We employ the LLM itself as a router to assess query difficulty before dispatching to the appropriate mode.
    \item \textbf{Speculative (Spec-TF)}: We operate via two mechanisms. \textbf{Spec-TF (Trigger)} constructs a model-specific uncertainty keyword library (e.g., \emph{wait}, \emph{hmm}) that varies across models, triggering a re-think during inference. \textbf{Spec-TF (Entropy)} monitors token-level output probabilities and triggers mode escalation when entropy exceeds a calibrated threshold.
    \end{itemize}

\item \textbf{Training-Based Implementation}: Built on MathLightEval \cite{hendrycksmath2021}, all training variants utilize a unified data construction pipeline based on Rejection Fine-Tuning (RFT) with multiple rollouts per problem:
    \begin{itemize}[nosep,leftmargin=*]
    \item \textbf{SFT}: We train on the sample that are both \emph{correct} and \emph{token-minimal} in multiple rollout results. For Prompt-Tuning (PT-SFT) and Speculative (Spec-SFT), the model is directly fine-tuned on these samples to autonomously select modes or trigger escalation. We choose the Spec-TF (Entropy) for Spec-SFT because it achieves a better performance. For Routing, the optimal mode serves as the ground-truth label to train either the LLM itself (RT-SFT).
    \item \textbf{DPO}: The RFT process naturally yields preference pairs. The chosen sample is the correct, token-minimal response. Rejected samples are longer correct answers, incorrect answers, or sub-optimal routing modes. This optimizes both prompt-tuning (PT-DPO), router (RT-DPO), and speculative (Spec-DPO).
    \item \textbf{GRPO}: In on-policy RL training, a unified reward structure is applied during rollouts to optimize autonomous mode selection (PT-GRPO), router policies (RT-GRPO), and speculative decoding triggers (Spec-GRPO).
    \end{itemize}
\end{itemize}

\paragraph{External methods.} We integrate 12 representative methods from the community into our unified pipeline, covering all three strategies:
\begin{itemize}[nosep,leftmargin=*]
\item \textbf{Prompt-Tuning}: S1 \citep{muennighoff2025s1}, TALE \citep{han2025tale}, Budget-Guidance \cite{li2025steering}, Sketch-of-Thought (SoT) \cite{aytes2025sketch}, Chain-of-Draft (CoD) \cite{xu2025chain}, DynaThink \cite{pan2024dynathink}, DEER \citep{yang2025dynamic} and RASC \citep{wan2025reasoning}.

\item \textbf{Routing}: AdaptThink \citep{zhang2025adaptthink} (GRPO-trained router) and HDFlow \citep{yao2024hdflow} (rule-based difficulty routing).

\item \textbf{Speculative}: MixReasoning \citep{lu2025mixreasoning} (entropy-based) and ADR \citep{zhang2025adr} (SFT+GRPO trained switching policy).
\end{itemize}

All external methods are re-implemented within our unified pipeline and evaluated under identical conditions for fair comparison. 
Reproduction details and any deviations from original papers are documented in Appendix~\ref{app:impl_details}.

\section{Effectiveness--Efficiency Trade-off of Switching Strategies}
\label{sec:tradeoff}

\textbf{RQ1:} \emph{How do different thinking-mode switch strategies (\PT{}/\RT{}/\Spec{}) trade off between effectiveness (accuracy) and efficiency (token cost)?}

To answer RQ1, we evaluate all three strategy implementations across all five benchmarks and six LLMs, examining how each strategy balances accuracy against token cost.

\begin{table*}[t]
\centering
\small
\setlength{\tabcolsep}{3.5pt}
\begin{tabular}{l|cc|cc|cc|cc|cc|cc}
\toprule
& \multicolumn{2}{c|}{\textbf{MATH500}} & \multicolumn{2}{c|}{\textbf{AIME 2025}} & \multicolumn{2}{c|}{\textbf{GPQA}} & \multicolumn{2}{c|}{\textbf{LiveCode}} & \multicolumn{2}{c|}{\textbf{Codeforces}} & \multicolumn{2}{c}{\textbf{AVG}} \\
\textbf{Method} & Acc & Tok & Acc & Tok & Acc & Tok & Acc & Tok & Acc & Tok & Acc & Tok \\
\midrule
\multicolumn{13}{l}{\emph{Baselines (Qwen3.5-9B)}} \\
\midrule
Full-Think & 85.0 & 9.1k & 70.0 & 23.1k & 54.5 & 15.9k & 21.6 & 16.1k & 23.5 & 17.6k & 40.9 & 16.4k \\
No-Think & \textbf{86.8} & \underline{1.7k} & 53.3 & \underline{12.8k} & 53.0 & \underline{3.1k} & 35.9 & \underline{5.0k} & 32.8 & \underline{4.8k} & 42.4 & \underline{5.5k} \\
\midrule
\multicolumn{13}{l}{\emph{Our Strategy Implementations (Training-Free)}} \\
\midrule
PT-TF & 85.4 & 6.5k & \textbf{80.0} & 20.2k & \textbf{63.1} & 12.2k & 29.9 & 11.7k & 29.5 & 11.2k & \textbf{47.6} & 12.4k \\
RT-TF & 85.8 & 6.2k & 73.3 & 22.4k & 52.0 & 11.1k & 29.9 & 15.1k & 29.2 & 16.8k & 44.1 & 14.3k \\
Spec-TF (Trigger) & 86.2 & 7.8k & 60.0 & 32.2k & 52.5 & 15.0k & 37.1 & 17.2k & 35.8 & 18.6k & 44.3 & 18.2k \\
Spec-TF (Entropy) & 84.8 & 11.1k & 63.3 & 32.7k & 55.6 & 18.4k & 43.1 & 20.8k & 32.2 & 23.0k & 45.8 & 21.2k \\
\bottomrule
\end{tabular}
\caption{Strategy-level trade-off on Qwen3.5-9B. \PT{} achieves the best accuracy with 24\% token reduction; \RT{} preserves accuracy with 13\% saving; \Spec{} boosts accuracy at extra token cost.}
\label{tab:main_results}
\end{table*}

\begin{figure}[t]
    \centering
    \includegraphics[width=\columnwidth]{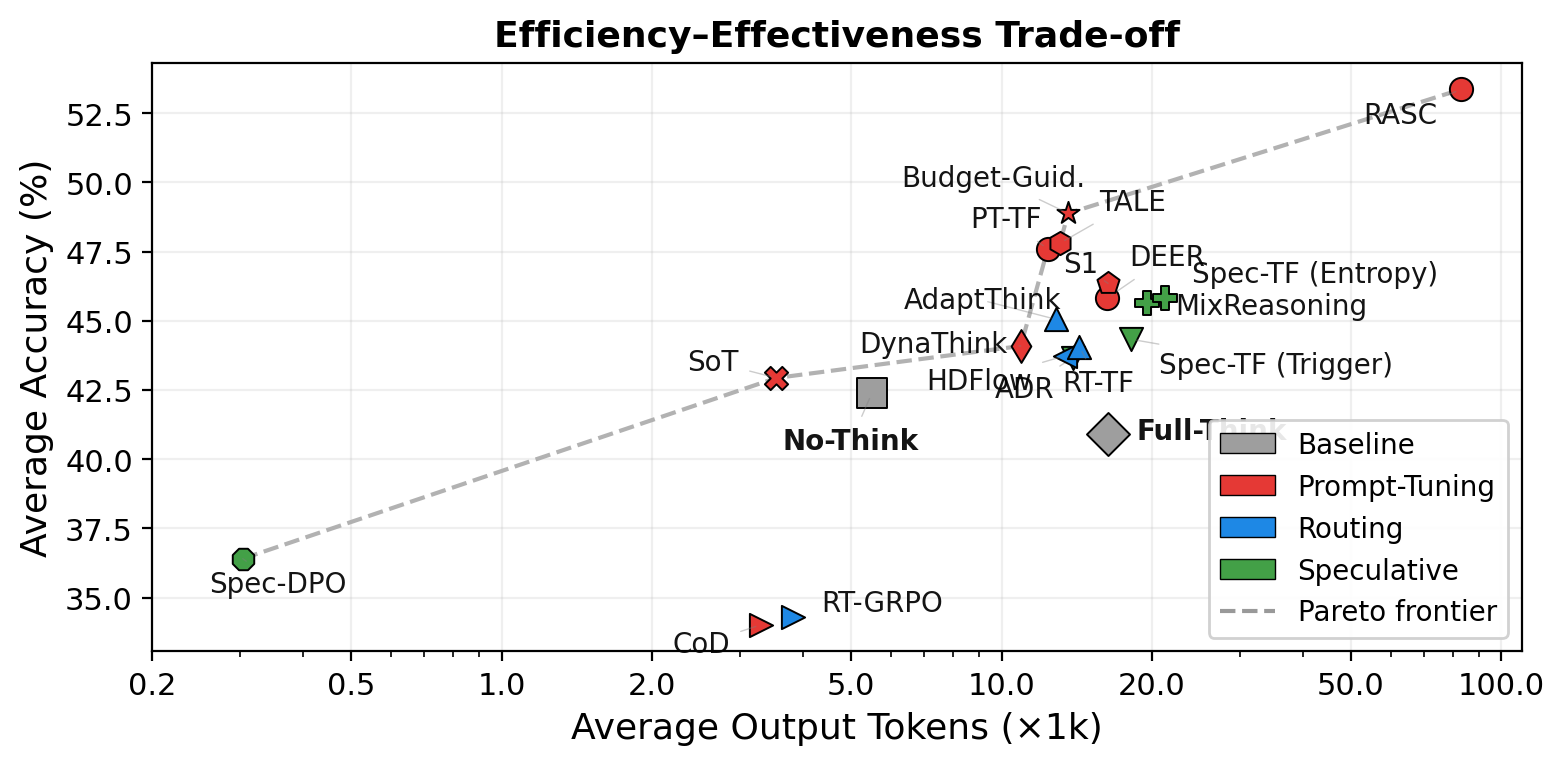}
    \caption{Efficiency--effectiveness trade-off on Qwen3.5-9B. Each point represents a method averaged over 5 datasets; dashed line shows the Pareto frontier.}
    \label{fig:pareto}

\end{figure}

\begin{table*}[t]
\centering
\small
\begin{tabular}{l|r|cc|cc|cc|cc|cc}
\toprule
& & \multicolumn{2}{c|}{\textbf{Full-Think}} & \multicolumn{2}{c|}{\textbf{No-Think}} & \multicolumn{2}{c|}{\textbf{PT-TF}} & \multicolumn{2}{c|}{\textbf{RT-TF}} & \multicolumn{2}{c}{\textbf{Spec-TF}} \\
\textbf{Model} & \textbf{Size} & Acc & Tok & Acc & Tok & Acc & Tok & Acc & Tok & Acc & Tok \\
\midrule
Qwen3.5-2B & 2B & 11.8 & 26.6k & 18.7 & 14.0k & 13.2 & 29.2k & 14.0 & 24.7k & \textbf{14.1} & 35.3k \\
Qwen3.5-9B & 9B & 40.9 & 16.4k & 42.4 & 5.5k & \textbf{47.6} & 12.4k & 44.1 & 14.3k & 45.8 & 21.2k \\
gpt-oss-20B & 20B & 36.4 & 19.7k & 24.5 & 0.8k & 32.9 & 4.0k & 29.2 & 9.1k & \textbf{36.8} & 20.4k \\
Seed-OSS-36B & 36B & 51.5 & 13.5k & 27.0 & 915 & \textbf{46.3} & 8.2k & 39.2 & 7.4k & 43.9 & 10.0k \\
DeepSeek-V3.1 & 671B & 67.4 & 3.2k & 63.7 & 1.7k & 74.7 & 3.3k & 68.9 & 2.7k & \textbf{75.8} & 4.2k \\
Kimi-K2.5 & 1.1T & 73.9 & 8.9k & 72.3 & 2.4k & \textbf{80.8} & 7.4k & 77.8 & 7.4k & 74.5 & 10.3k \\
\bottomrule
\end{tabular}
\caption{Model scale modulation: effectiveness and efficiency of fixed baselines and three Training-Free strategies across 6 models (2B--1.1T), averaged over five benchmarks. Spec-TF reports the Entropy variant.}
\label{tab:scale}
\end{table*}

\subsection{Overall Trade-off Patterns}
\label{sec:tradeoff_overall}


Table~\ref{tab:main_results} and Figure~\ref{fig:pareto} reveal that the \emph{three strategies exhibit fundamentally different trade-off patterns between effectiveness and efficiency}:

\paragraph{Finding 1: \PT{} consistently achieves Pareto-optimal trade-offs.} As shown in Figure~\ref{fig:pareto}, PT-TF simultaneously improves accuracy over Full-Think while substantially reducing token cost. This ``win-win'' pattern is unique to Prompt-Tuning: the prompt guides the model to allocate reasoning effort proportionally to difficulty, thereby avoiding unnecessary reasoning on simpler problems. 
Across all PT implementations in our benchmark, this Pareto-dominant behavior holds robustly.

\paragraph{Finding 2: \RT{} preserves effectiveness with moderate efficiency gains.} RT-TF maintains accuracy comparable to Full-Think while achieving moderate token savings through selective routing. The router correctly identifies easier problems (e.g., $\sim$60\% of MATH500) and routes them to \nothink{} mode, while conservatively keeping harder benchmarks in full reasoning mode. This conservative strategy yields steady but limited improvements.

\paragraph{Finding 3: \Spec{} boosts accuracy at extra token cost.} Unlike \PT{} and \RT{}, Spec-TF increases token usage relative to Full-Think, but in return yields notable accuracy improvements, particularly on code tasks where the ``try-then-verify'' mechanism excels. The no-think initial pass catches easy problems efficiently, but re-triggering deep reasoning when uncertainty is detected adds overhead. \Spec{} thus functions as an effectiveness-enhancing rather than efficiency strategy.

\begin{figure}[t]
    \centering
    \includegraphics[width=\columnwidth]{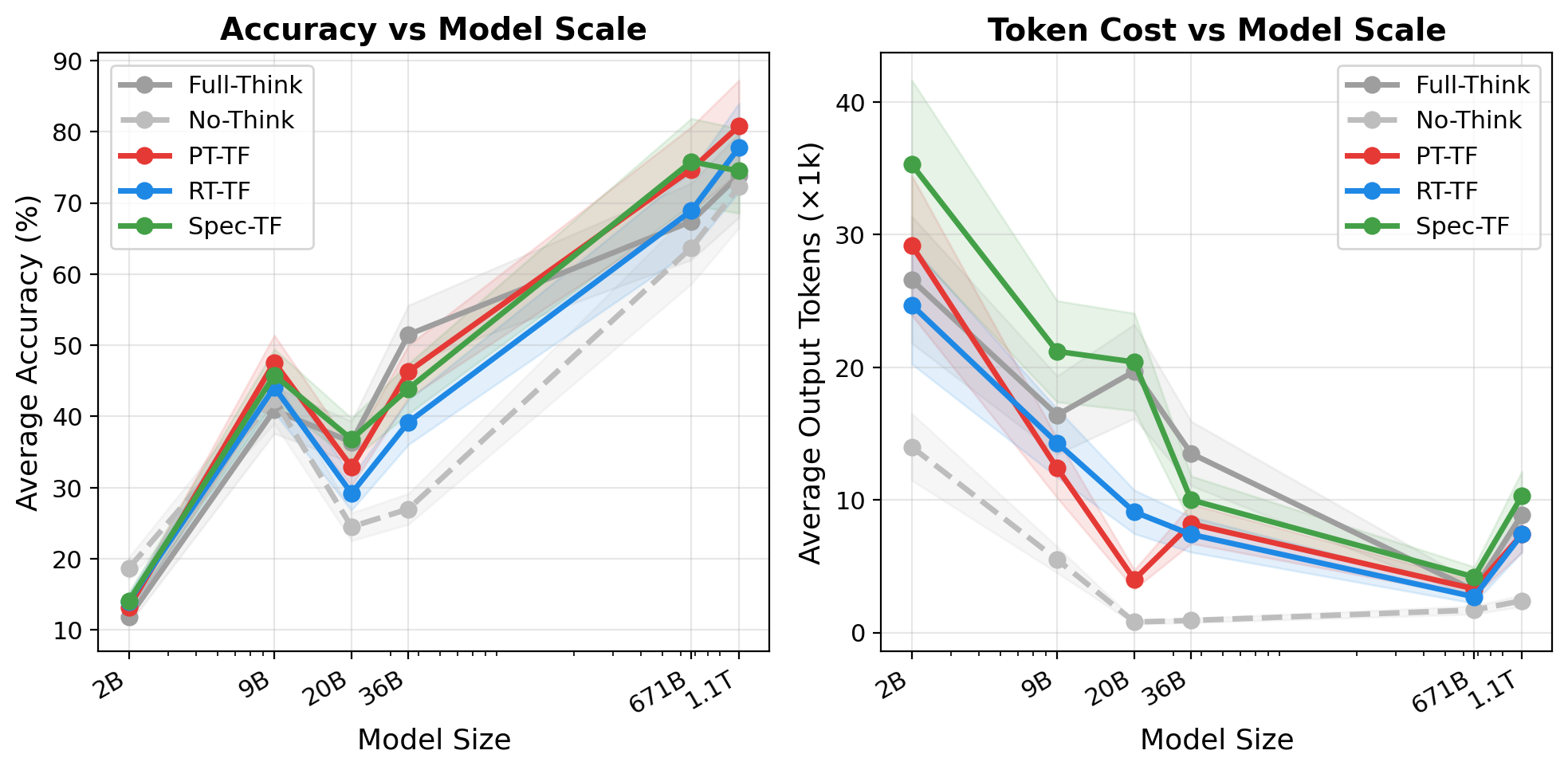}
    \caption{Strategy effectiveness (left) and efficiency (right) across model scales (2B--1.1T). All three strategies are evaluated across 6 models.}
    \label{fig:scale_trend}
    \vspace{-10pt}
\end{figure}

\subsection{Model Scale Effect}
\label{sec:tradeoff_scale}
To validate that trade-off patterns shift with model scale, we evaluate all six models (2B--1.1T) under Training-Free configurations. 
Table~\ref{tab:scale} reports averaged results, and Figure~\ref{fig:scale_trend} visualizes the strategy ranking evolution across scales.



\begin{figure*}[t]
    \centering
    \includegraphics[width=\textwidth]{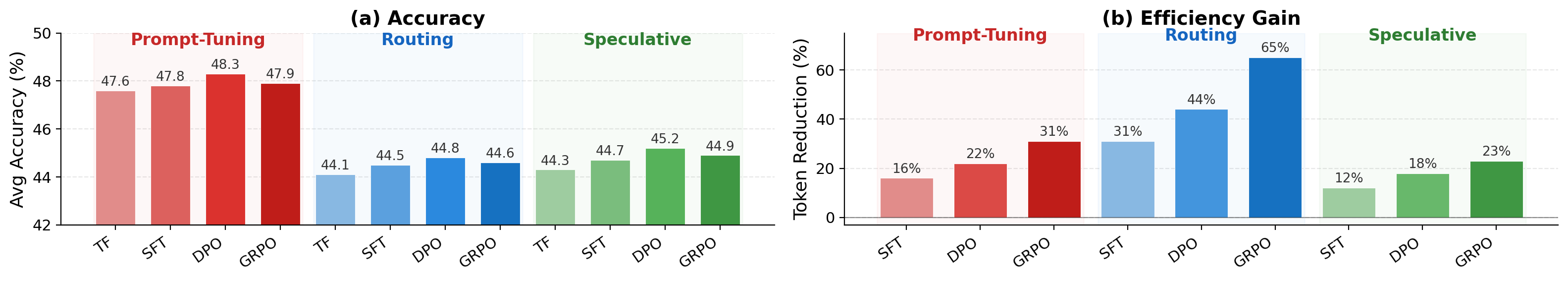}
    \caption{Training effect on switching capacity (Qwen3.5-9B). (a)~Accuracy across training regimes, averaged over five benchmarks. (b)~Token reduction relative to TF baseline.}
    \label{fig:training_effect}
\end{figure*}
We observe that the \emph{effectiveness--efficiency trade-off of each strategy shifts substantially with model scale}---neither strategy ranking nor efficiency advantage is consistent across scales:

\paragraph{Finding 4: Strategy effectiveness ranking varies across model scales.} 
The best strategy choice differs depending on the size (Table~\ref{tab:scale}): at 9B and 1.1T, \PT{} leads (47.6\% and 80.8\% respectively); at 20B and 671B, \Spec{} overtakes (36.8\% vs.\ 32.9\% at 20B; 75.8\% vs.\ 74.7\% at 671B); while at 2B, all three strategies perform similarly (13.2--14.1\%). \RT{} generally ranks last but remains competitive at larger scales (77.8\% at 1.1T). 
This scale-dependent ranking suggests that no single strategy universally dominates in effectiveness.

\paragraph{Finding 5: Strategy efficiency ranking is also scale-dependent.} Token efficiency does not uniformly favor one strategy across scales. Notably, \PT{} increases token usage at 2B (29.2k vs.\ 26.6k for Full-Think), while achieving strong savings at 36B ($-$39\%) and 1.1T ($-$17\%). In contrast, \RT{} is the most consistent in reducing token cost: it achieves savings at every scale from 9B onward (e.g., $-$13\% at 9B, $-$45\% at 36B, $-$17\% at 1.1T). \Spec{} consistently incurs extra tokens across all scales due to its re-think mechanism. These patterns indicate that efficiency-oriented deployment must carefully match the strategy to the target model scale.

\begin{table}[t]
\centering
\small
\setlength{\tabcolsep}{3pt}
\resizebox{\columnwidth}{!}{%
\begin{tabular}{l|cc|cc|cc}
\toprule
& \multicolumn{2}{c|}{\textbf{Math}} & \multicolumn{2}{c|}{\textbf{Science}} & \multicolumn{2}{c}{\textbf{Code}} \\
\textbf{Method} & $\Delta$Acc & Red\% & $\Delta$Acc & Red\% & $\Delta$Acc & Red\% \\
\midrule
No-Think & $-$7.4 & +62.8 & $-$1.5 & +80.6 & \textbf{+11.8} & +70.6 \\
\midrule
PT-TF & +5.2 & +20.5 & \textbf{+8.6} & +23.2 & +7.2 & +31.8 \\
RT-TF & +2.1 & +17.4 & $-$2.5 & +30.4 & +7.1 & +5.4 \\
Spec-TF & $-$4.4 & $-$12.4 & $-$2.0 & +5.3 & \textbf{+13.9} & $-$6.4 \\
\bottomrule
\end{tabular}%
}
\caption{Domain modulation on Qwen3.5-9B. $\Delta$Acc: accuracy change (percentage points) vs.\ Full-Think; Red\%: token reduction vs.\ Full-Think.}
\label{tab:domain}
\end{table}

\subsection{Task Domain Effect}
\label{sec:tradeoff_domain}

We further analyze how trade-off patterns vary across three task domains: math, science, and coding tasks. Table~\ref{tab:domain} reveals striking domain-dependent strategy preferences, demonstrating that the underlying nature of the task influences strategy selection:

\paragraph{Finding 6: The optimal strategy differs across task domains.}
No single strategy universally dominates: in Math and Science, \PT{} is the clear winner, improving both accuracy and token efficiency; in Code, however, \Spec{} achieves the largest accuracy boost via its try-then-verify'' mechanism, though \PT{} and \RT{} also yield efficient gains. 
This domain-dependent variation provides motivation for adaptive mode switching.

\subsection{Summary}
Overall, these domain-dependent patterns (\S\ref{sec:tradeoff_domain}), combined with model scale modulation (\S\ref{sec:tradeoff_scale}), confirm that no single strategy dominates universally. 
Consequently, an appropriate thinking mode switching strategy should carefully account for both the model scale and the expected task domain.

\section{Effect of Training Pipeline on Switching Strategies}
\label{sec:training}

\textbf{RQ2:} \emph{How do different training regimes (e.g., SFT/DPO/GRPO) affect the three thinking mode switch strategies?}

To answer RQ2, we train Qwen3.5-9B under three regimes (i.e., SFT, DPO, and GRPO) applied to each of the three strategies, and compare against the Training-Free (TF) baselines from \S\ref{sec:tradeoff}. 
All training experiments use MathLightEval as the training data source. 
Figure~\ref{fig:training_effect} summarizes the accuracy and efficiency results across all 5 benchmarks.

\begin{figure*}[t]
    \centering
    \includegraphics[width=\textwidth]{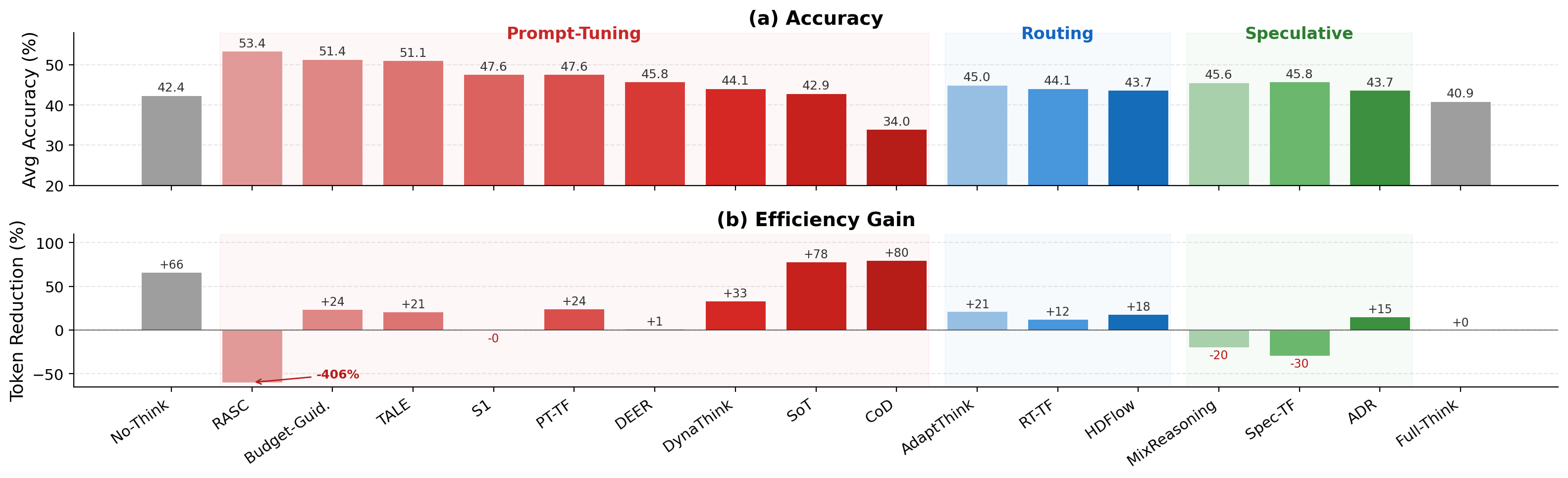}
    \caption{Fair comparison of 12 methods under unified evaluation on Qwen3.5-9B. (a)~Average accuracy across five benchmarks. (b)~Token reduction relative to Full-Think (positive = saving). Methods grouped by strategy.}
    \label{fig:fair_comparison}
\end{figure*}

\paragraph{Finding 7: Training universally improves switching capacity, with larger gains in efficiency than accuracy.} Across all three strategies, training (SFT, DPO, and GRPO) maintains or slightly improves accuracy compared to TF, while achieving substantially larger gains in token reduction (Figure~\ref{fig:training_effect}). This indicates that training primarily teaches the model when to skip unnecessary reasoning, rather than improving the reasoning itself. The accuracy improvements are modest (within 1-2 percentage points of TF), whereas efficiency gains range from 12\% to 65\% depending on the strategy and training method.

\paragraph{Finding 8: \RT{} benefits the most from training in efficiency.} The efficiency gains from training are strongly strategy-dependent: GRPO achieves 65\% token reduction for \RT{}, compared to 31\% for \PT{} and 23\% for \Spec{}. 
This disparity arises because \RT{}'s binary routing decision is well-matched to the training signal---correctness feedback directly reflects routing quality, enabling the router to learn a sharper difficulty boundary. 
In contrast, \PT{} and \Spec{} involve more fine-grained decisions (reasoning depth and trigger timing, respectively) that are harder to optimize with coarse reward signals.

\paragraph{Finding 9: Different training methods offer distinct trade-off profiles.} 
The three training regimes serve a complementary role.
DPO yields the largest accuracy improvements and is optimal for improving effectiveness. 
Conversely, GRPO achieves the greatest efficiency gains and token savings, making it ideal when efficiency is the priority. 
Meanwhile, SFT provides a balanced middle ground in both dimensions, serving as a safe default when specific optimization targets are unclear.

\section{Fair Comparison of Existing Methods}
\label{sec:fair_comparison}

\textbf{RQ3:} \emph{Under a unified pipeline (e.g., same model and data), do the advantages claimed by existing methods hold, and are the conclusions consistent with our strategy-level findings?}

The RQ3 also aligns with the core motivation of \ours{}: existing methods are developed and evaluated in isolation, making cross-method comparison unreliable. 
We re-implement 12 representative external methods within our unified pipeline and compare them against our TF baselines (PT-TF, RT-TF, Spec-TF) under identical conditions. 
Figure~\ref{fig:fair_comparison} visualizes the results, and full per-dataset breakdowns are in Table~\ref{tab:fair_comparison} (Appendix).

\paragraph{Finding 10: External methods exhibit similar strategy-level trade-off patterns.} The unified evaluation validates our findings from \S\ref{sec:tradeoff}:
\begin{itemize}[nosep,leftmargin=*]
    \item \textbf{Prompt-Tuning} methods span the widest Pareto frontier, ranging from RASC (53.4\% accuracy, 5$\times$ more tokens) to CoD (34.0\%, +80\% token reduction), confirming \PT{}'s unique ability to deliver diverse operating points along the efficiency--effectiveness spectrum. Our PT-TF baseline (47.6\%, +24\% token reduction) sits at a competitive position within this range.
    \item \textbf{Routing} based methods consistently show moderate token savings (18--21\%) with preserved accuracy (43--45\%), consistent with \RT{}'s pattern of maintaining effectiveness while gaining modest efficiency. Our RT-TF (44.1\%, +12.5\%) also aligns with this trend.
    \item \textbf{Speculative} based methods mostly show negative token reduction ($-$11\% to $-$30\%), confirming their role as accuracy-boosting rather than efficiency-oriented approaches, which aligns with our analysis in \S\ref{sec:tradeoff_overall}. ADR is the only external Spec method achieving token reduction. 
\end{itemize}

\paragraph{Finding 11: No single external method dominates across all task domains.} As shown in Table~\ref{tab:fair_comparison}, top methods vary by domain: RASC excels on AIME (83.3\%), Budget-Guidance leads on MATH500 (87.8\%), and SoT achieves the best GPQA accuracy.
Overall, this also reinforces that practitioners should select suitable methods based on their target task domain rather than relying on the reported aggregate score.

\section{Conclusion}
\label{sec:conclusion}

In this paper, we proposed \ours{}, a unified benchmark for systematically understanding thinking-mode switch strategies in hybrid-reasoning LLMs. By evaluating 12 configurations (3 strategies $\times$ 4 training regimes) across 6 models and 5 datasets, and integrating 12+ external methods into the same pipeline, we reveal that: (1) the three strategies exhibit fundamentally different trade-off profiles; (2) training gains are strongly strategy-dependent; and (3) the optimal strategy--training combination shifts with model scale and task domain. We release all implementations and the unified evaluation framework to facilitate future research on efficient hybrid reasoning.

\section*{Limitations}

Our work has several limitations. First, training-based evaluations are limited to Qwen3.5-9B due to computational constraints, scaling to 20B+ models would strengthen conclusions about training-scale interactions.
Second, our evaluation focuses on single-turn reasoning; multi-turn and agentic scenarios where mode-switching decisions compound across steps remain unexplored.
Third, while we cover mathematics, science, and code, other domains (e.g., creative writing, multilingual tasks) may exhibit different trade-off patterns.

\section*{Ethical Statement}
We use LLMs for paper polishing and figure plotting, and have carefully verified all outputs for correctness.
This work evaluates existing LLMs on publicly available benchmarks and does not involve human subjects, private data, or the generation of harmful content. All datasets used (MATH500, AIME 2025, GPQA-Diamond, LiveCodeBench, Codeforces) are publicly released for research purposes. Our benchmark focuses on improving inference efficiency of reasoning models, which may reduce computational costs and associated carbon emissions. We do not foresee direct negative societal impacts from this work. All model outputs are used solely for automated evaluation and are not deployed in user-facing applications.


\bibliography{references}

\appendix

\clearpage

\section{Prompt Templates}
\label{app:prompts}

This appendix provides the system prompts and user templates used by each strategy and model family. All prompts are centralized to ensure consistency between training data construction and inference. We reproduce them in full below.

\subsection{Answer Format Instructions}
\label{app:answer_format}

Every user message in our benchmark is appended with a domain-specific answer format instruction:

\paragraph{Mathematics (MATH500, AIME).}
\begin{quote}
\small\ttfamily
Put your final answer within \textbackslash boxed\{\}.
\end{quote}

\paragraph{Code (LiveCodeBench, Codeforces).}
\begin{quote}
\small\ttfamily
Write a Python solution. Read input from stdin and print output to stdout. Do not include any test code or examples. Only provide the solution code.
\end{quote}

\paragraph{Science (GPQA multiple-choice).}
\begin{quote}
\small\ttfamily
Provide your answer. If the problem is multiple-choice, state the correct option letter. Otherwise, provide a clear, concise answer.
\end{quote}

\subsection{Problem Templates}
\label{app:problem_templates}

The user-facing message is constructed by formatting the problem text into domain-specific templates:

\paragraph{Math user message.}
\begin{quote}
\small\ttfamily
\{problem\}\\[3pt]
Put your final answer within \textbackslash boxed\{\}.
\end{quote}

\paragraph{Code user message.}
\begin{quote}
\small\ttfamily
\{problem\}\\[3pt]
Write a Python solution. Read input from stdin and print output to stdout. Do not include any test code or examples. Only provide the solution code.
\end{quote}

\paragraph{Science user message.}
\begin{quote}
\small\ttfamily
\{problem\}\\[3pt]
Provide your answer. If the problem is multiple-choice, state the correct option letter. Otherwise, provide a clear, concise answer.
\end{quote}

These templates are shared by all baselines (Full-Think, No-Think) and strategies. Strategies add their own system prompts (below) while keeping the user message format unchanged.

\subsection{Prompt-Tuning System Prompts}
\label{app:pt_prompts}

PT-TF uses model-specific system prompts that teach the model to self-select reasoning depth using its native thinking-mode interface. The complete prompts are reproduced verbatim below.

\paragraph{Qwen3.5 series (\texttt{PROMPT\_TUNING\_SYSTEM\_QWEN}).}
\begin{quote}
\small\ttfamily
You are an expert problem solver with adaptive reasoning.\\[3pt]
Before solving, assess the problem's difficulty and choose your reasoning depth:\\
- For simple problems: Keep your <think> block empty or very brief. Answer directly.\\
- For complex problems: Use your <think> block for thorough step-by-step reasoning.\\
- For medium problems: Use your <think> block briefly for key observations only.\\[3pt]
You decide the appropriate depth based on the problem.
\end{quote}

\paragraph{gpt-oss (\texttt{PROMPT\_TUNING\_SYSTEM\_GPT\_OSS}).}
\begin{quote}
\small\ttfamily
You are an expert problem solver. You may adjust your reasoning level between high, medium, and low based on problem complexity.\\[3pt]
- For simple problems: Use low reasoning effort. Minimal analysis, direct answer.\\
- For complex problems: Use high reasoning effort. Thorough step-by-step analysis.\\
- For medium problems: Use medium reasoning effort. Brief analysis on key steps.\\[3pt]
Assess each problem and choose the appropriate reasoning level yourself.
\end{quote}

\paragraph{Seed-OSS (\texttt{PROMPT\_TUNING\_SYSTEM\_SEED\_OSS}).}
\begin{quote}
\small\ttfamily
You are an intelligent assistant with reflective reasoning ability. You may adjust your thinking budget based on problem complexity.\\[3pt]
- For simple problems: Set your thinking budget to 0 or minimal. Skip the thinking process and answer directly.\\
- For complex problems: Allow a generous thinking budget (4096+ tokens). Think thoroughly and use <seed:cot\_budget\_reflect> to track your token usage.\\
- For medium problems: Set a moderate thinking budget (512-1024 tokens). Think on key steps, reflect on progress, then answer.\\[3pt]
Example reflection during thinking:\\
<seed:cot\_budget\_reflect>I have used 200 tokens, and there are 800 tokens remaining for use.</seed:cot\_budget\_reflect>\\[3pt]
Assess each problem and manage your thinking budget accordingly.
\end{quote}

\paragraph{PT-TF user template (\texttt{PROMPT\_TUNING\_USER\_TEMPLATE}).} This wraps the problem before appending the answer format instruction:
\begin{quote}
\small\ttfamily
Solve the following problem. Decide whether it requires deep thinking or a direct answer.\\[3pt]
Problem: \{problem\}\\[3pt]
Provide your final answer.
\end{quote}

The complete user message sent to the model is the combination of user messages and tailored system prompt.

\subsection{Routing Prompts}
\label{app:rt_prompts}

RT-TF uses a two-stage process. Each stage uses different prompts.

\subsubsection{Stage 1: Judge Prompts}

The judge stage uses model-specific prompts to classify problem difficulty. It runs in no-think/low-effort mode with \texttt{max\_tokens=256}.

\paragraph{Qwen3.5 judge (\texttt{ROUTING\_JUDGE\_SYSTEM\_QWEN} + \texttt{ROUTING\_JUDGE\_USER\_QWEN}).}
\begin{quote}
\small\ttfamily
\textbf{[System]} You are a problem difficulty classifier.\\[6pt]
\textbf{[User]} Assess the difficulty of the following problem and choose a reasoning mode.\\[3pt]
Problem: \{problem\}\\[3pt]
Available modes:\\
1 - Think: Complex problem, enable full thinking with <think> block\\
2 - NoThink: Simple problem, skip thinking, answer directly\\
3 - Budget Think: Medium problem, think within a limited token budget\\[3pt]
Respond with ONLY a JSON object:\\
\{"mode": "1" or "2" or "3", "budget": null or 1024 or 2048 or 4096\}\\[3pt]
Rules:\\
- Mode 1: budget = null (unlimited thinking)\\
- Mode 2: budget = null (no thinking)\\
- Mode 3: budget = 1024 / 2048 / 4096
\end{quote}

\paragraph{gpt-oss judge (\texttt{ROUTING\_JUDGE\_SYSTEM\_GPT\_OSS} + \texttt{ROUTING\_JUDGE\_USER\_GPT\_OSS}).}
\begin{quote}
\small\ttfamily
\textbf{[System]} You are a problem difficulty classifier.\\[6pt]
\textbf{[User]} Assess the difficulty of the following problem and choose a reasoning effort level.\\[3pt]
Problem: \{problem\}\\[3pt]
Available reasoning levels:\\
1 - High: Complex problem, needs thorough step-by-step analysis\\
2 - Low: Simple problem, minimal analysis, direct answer\\
3 - Medium: Moderate problem, brief analysis on key steps\\[3pt]
Respond with ONLY a JSON object: \{"level": "high" or "medium" or "low"\}
\end{quote}

\paragraph{Seed-OSS judge (\texttt{ROUTING\_JUDGE\_SYSTEM\_SEED\_OSS} + \texttt{ROUTING\_JUDGE\_USER\_SEED\_OSS}).}
\begin{quote}
\small\ttfamily
\textbf{[System]} You are a problem difficulty classifier.\\[6pt]
\textbf{[User]} Assess the difficulty of the following problem and choose a thinking strategy.\\[3pt]
Problem: \{problem\}\\[3pt]
Available thinking modes:\\
1 - Full Think: Complex problem, unlimited thinking budget\\
2 - No Think: Simple problem, skip thinking (budget = 0)\\
3 - Budget Think: Medium problem, think within a fixed token budget\\[3pt]
Respond with ONLY a JSON object:\\
\{"mode": "1" or "2" or "3", "budget": null or 512 or 1024 or 2048 or 4096\}\\[3pt]
Rules:\\
- Mode 1: budget = null (unlimited)\\
- Mode 2: budget = null (no thinking)\\
- Mode 3: budget = 512 / 1024 / 2048 / 4096
\end{quote}

\subsubsection{Stage 2: Solve Prompt}

After routing, the problem is solved using a shared system prompt (\texttt{ROUTING\_SOLVE\_SYSTEM}):
\begin{quote}
\small\ttfamily
You are a helpful assistant. Solve the given problem carefully.
\end{quote}
The user message uses the standard problem template (\S\ref{app:problem_templates}).

\subsection{Speculative Thinking Configuration}
\label{app:spec_prompts}

Speculative strategies do not use custom system prompts; they reuse the baseline message format (\S\ref{app:problem_templates}). Mode switching is controlled at the \emph{token level} via two-pass generation.

\paragraph{Spec-Trigger: Complete keyword library (\texttt{SPECULATIVE\_TRIGGER\_WORDS}).} The full 55-keyword library:
\begin{itemize}[nosep,leftmargin=*]
\item \textbf{Hesitation/Uncertainty (9):} \emph{wait, hmm, hm{,}, hold on, i'm not sure, i am not sure, not certain, unclear, confusing}
\item \textbf{Self-correction/Backtracking (14):} \emph{actually, on second thought, let me reconsider, i made a mistake, i made an error, that's wrong, that's incorrect, that doesn't seem right, this is wrong, correction:, i need to correct, scratch that, let me redo, start over, going back}
\item \textbf{Re-examination/Verification (11):} \emph{let me verify, let me check, let me re-examine, let me recalculate, double-check, double check, verify this, verify that, reconsider, re-examine, revisit}
\item \textbf{Alternative approach (10):} \emph{alternatively, another approach, another way, different approach, different method, try a different, let me try, instead{,}, perhaps, maybe i should}
\item \textbf{Deeper reasoning (11):} \emph{think again, think more carefully, think step by step, let me think, need to think, this requires, this is tricky, this is complex, this is harder, more carefully, closer look}
\item \textbf{Contradiction/Confusion (9):} \emph{but that contradicts, that contradicts, this contradicts, doesn't make sense, does not make sense, something is off, something is wrong, paradox, inconsistent}
\item \textbf{Explicit re-reasoning (7):} \emph{recap, summarize what we know, let me summarize, to be more precise, more precisely, to clarify, in other words}
\end{itemize}
All matching is case-insensitive. If \emph{any} keyword appears in the no-think output, the response is discarded and regenerated in full think mode.

\paragraph{Spec-Entropy: Threshold configuration.} Model-specific entropy thresholds (calibrated on held-out validation set): Qwen3.5 $\tau{=}0.10$, gpt-oss $\tau{=}0.08$, Seed-OSS $\tau{=}0.06$. All use top-$k{=}20$ logprobs. Escalation is triggered when $\geq 3$ tokens or $>5\%$ of output tokens exceed the threshold.

Entropy is computed as normalized Shannon entropy from the top-$k$ logprobs:
\begin{equation}
H_t = \frac{-\sum_{v \in \text{top-}k} \hat{p}_t(v) \log \hat{p}_t(v)}{\log k}
\end{equation}
where $\hat{p}_t(v) = \exp(\ell_v) / \sum_{v'} \exp(\ell_{v'})$ is the renormalized probability over the top-$k$ returned logprobs. Lower thresholds for gpt-oss/Seed-OSS reflect their generally tighter output distributions in no-think mode.

\subsection{Training Prompts}
\label{app:training_prompts}

\paragraph{Baseline SFT mode-selection (\texttt{SFT\_MODE\_SELECTION\_SYSTEM}).}
\begin{quote}
\small\ttfamily
You are an adaptive reasoning assistant. For each problem, you must first decide your reasoning strategy, then solve the problem accordingly.\\[3pt]
Output format:\\
{[}MODE: think{]} or {[}MODE: nothink{]}\\
Then solve the problem.
\end{quote}

\paragraph{Strategy-specific training prompts.} The function \texttt{get\_strategy\_system\_prompt(strategy, model\_family)} returns the \emph{same} prompt used at inference time:
\begin{itemize}[nosep,leftmargin=*]
\item \texttt{strategy="pt"}: Returns the PT system prompt for the given model family (\S\ref{app:pt_prompts})
\item \texttt{strategy="rt"}: Returns \texttt{ROUTING\_SOLVE\_SYSTEM} = ``You are a helpful assistant. Solve the given problem carefully.''
\item \texttt{strategy="baseline"}: Returns ``You are a helpful math assistant. Solve the problem step by step and provide your final answer.''
\end{itemize}
This design guarantees prompt consistency between training data construction and evaluation.

\subsection{Evaluation: LLM-as-Judge Prompt}
\label{app:judge_prompt}

For problems where rule-based answer extraction is insufficient, we use an LLM-as-judge for correctness evaluation:

\begin{quote}
\small\ttfamily
\textbf{[System]} You are an expert evaluator. Given a question, a reference answer, and a student's answer, determine if the student's answer is correct.\\[6pt]
\textbf{[User]} Question:\\
\{problem\}\\[3pt]
Reference Answer:\\
\{reference\}\\[3pt]
Student's Answer:\\
\{response\}\\[3pt]
Is the student's answer correct? Consider mathematical equivalence (e.g., 1/2 and 0.5 are equivalent, different forms of the same expression are equivalent).\\[3pt]
Respond with ONLY a JSON object: \{"correct": true\} or \{"correct": false\}
\end{quote}

\section{Full Per-Dataset Results}
\label{app:full_results}

\subsection{Fair Comparison Per-Dataset Results}

Table~\ref{tab:fair_comparison} provides the complete per-dataset breakdown of Figure~\ref{fig:fair_comparison}.

\begin{table*}[t]
\centering
\small
\setlength{\tabcolsep}{3.5pt}
\begin{tabular}{ll|cc|cc|cc|cc|cc|cc}
\toprule
& & \multicolumn{2}{c|}{\textbf{MATH500}} & \multicolumn{2}{c|}{\textbf{AIME 2025}} & \multicolumn{2}{c|}{\textbf{GPQA}} & \multicolumn{2}{c|}{\textbf{LiveCode}} & \multicolumn{2}{c|}{\textbf{Codeforces}} & \multicolumn{2}{c}{\textbf{AVG}} \\
\textbf{Strategy} & \textbf{Method} & Acc & Tok & Acc & Tok & Acc & Tok & Acc & Tok & Acc & Tok & Acc & Red\% \\
\midrule
\multicolumn{14}{l}{\emph{Fixed Baselines}} \\
\midrule
& Full-Think & 85.0 & 9.1k & 70.0 & 23.1k & 54.5 & 15.9k & 21.6 & 16.1k & 23.5 & 17.6k & 40.9 & --- \\
& No-Think & 86.8 & 1.7k & 53.3 & 12.8k & 53.0 & 3.1k & 35.9 & 5.0k & 32.8 & 4.8k & 42.4 & +66.5 \\
\midrule
\multicolumn{14}{l}{\emph{Prompt-Tuning}} \\
\midrule
PT & RASC & \textbf{89.2} & 42.5k & \textbf{83.3} & 131.7k & 56.6 & 89.1k & \textbf{49.7} & 79.0k & \textbf{38.0} & 72.5k & \textbf{53.4} & $-$406 \\
PT & Budget-Guidance (M) & 87.8 & 6.5k & 73.3 & 19.3k & 65.2 & 10.0k & 45.5 & 13.5k & 35.0 & 12.9k & 51.4 & +23.9 \\
PT & TALE (EP) & 87.2 & 6.8k & 76.7 & 21.6k & 67.2 & 10.3k & 43.7 & 14.3k & 30.6 & 12.0k & 51.1 & +21.2 \\
PT & S1 (High) & 84.4 & 16.3k & 70.0 & 26.2k & 54.5 & 16.0k & 43.7 & 17.0k & 35.2 & 18.0k & 47.6 & $-$0.4 \\
PT & PT-TF & 85.4 & 6.5k & 80.0 & 20.2k & 63.1 & 12.2k & 29.9 & 11.7k & 29.5 & 11.2k & 47.6 & +24.4 \\
PT & DEER & 85.6 & 8.3k & 66.7 & 26.2k & 52.5 & 16.8k & 37.1 & 15.9k & 37.2 & 13.9k & 45.8 & +1.1 \\
PT & DynaThink & 86.8 & 2.4k & 56.7 & 15.5k & 50.5 & 7.6k & 40.1 & 14.6k & 36.3 & 14.5k & 44.1 & +33.4 \\
PT & Sketch-of-Thought & 82.0 & 895 & 43.3 & 3.8k & \textbf{69.7} & 3.2k & 37.1 & 5.4k & 32.5 & 4.4k & 42.9 & +78.3 \\
PT & Chain-of-Draft & 73.8 & 854 & 26.7 & 4.5k & 65.7 & 3.6k & 29.9 & 2.9k & 24.0 & 3.7k & 34.0 & +79.9 \\
\midrule
\multicolumn{14}{l}{\emph{Routing}} \\
\midrule
RT & AdaptThink & 86.4 & 5.8k & 73.3 & 20.8k & 54.0 & 10.5k & 31.1 & 13.2k & 30.3 & 14.0k & 45.0 & +21.3 \\
RT & RT-TF & 85.8 & 6.2k & 73.3 & 22.4k & 52.0 & 11.1k & 29.9 & 15.1k & 29.2 & 16.8k & 44.1 & +12.5 \\
RT & HDFlow & 85.6 & 5.9k & 70.0 & 21.5k & 53.0 & 11.5k & 30.5 & 13.8k & 29.5 & 14.2k & 43.7 & +18.2 \\
RT & RT-GRPO & 85.8 & 2.1k & 73.3 & 7.1k & 54.0 & 3.9k & 30.5 & 5.8k & 29.4 & 6.2k & 44.6 & +69.5 \\
\midrule
\multicolumn{14}{l}{\emph{Speculative}} \\
\midrule
Spec & Spec-TF (Entropy) & 84.8 & 11.1k & 63.3 & 32.7k & 55.6 & 18.4k & 43.1 & 20.8k & 32.2 & 23.0k & 45.8 & $-$29.6 \\
Spec & MixReasoning & 85.4 & 9.8k & 63.3 & 30.5k & 54.5 & 17.2k & 41.3 & 19.5k & 33.6 & 20.8k & 45.6 & $-$19.6 \\
Spec & Spec-TF (Trigger) & 86.2 & 7.8k & 60.0 & 32.2k & 52.5 & 15.0k & 37.1 & 17.2k & 35.8 & 18.6k & 44.3 & $-$11.1 \\
Spec & ADR & 84.8 & 7.2k & 63.3 & 22.5k & 54.0 & 12.2k & 34.7 & 13.5k & 31.4 & 14.0k & 43.7 & +15.2 \\
Spec & Spec-DPO & 86.4 & 6.2k & 63.3 & 26.5k & 53.5 & 12.8k & 37.7 & 14.1k & 35.2 & 14.8k & 45.2 & +9.1 \\
\bottomrule
\end{tabular}
\caption{Fair comparison of methods under unified evaluation on Qwen3.5-9B. All methods use an identical model, datasets, and evaluation pipeline. Red\%: token reduction relative to Full-Think (positive = saving).}
\label{tab:fair_comparison}
\vspace{-10pt}
\end{table*}

\subsection{Training Effect Per-Dataset Results}

Table~\ref{tab:training_full} reports the complete per-dataset training results visualized in Figure~\ref{fig:training_effect}.

\begin{table*}[t]
\centering
\footnotesize
\setlength{\tabcolsep}{3pt}
\begin{tabular}{ll|cc|cc|cc|cc|cc|cc}
\toprule
& & \multicolumn{2}{c|}{\textbf{MATH500}} & \multicolumn{2}{c|}{\textbf{AIME 2025}} & \multicolumn{2}{c|}{\textbf{GPQA}} & \multicolumn{2}{c|}{\textbf{LiveCode}} & \multicolumn{2}{c|}{\textbf{Codeforces}} & \multicolumn{2}{c}{\textbf{AVG}} \\
\textbf{Para.} & \textbf{Train} & Acc & Tok & Acc & Tok & Acc & Tok & Acc & Tok & Acc & Tok & Acc & Tok \\
\midrule
\multicolumn{14}{l}{\emph{Baselines}} \\
\midrule
-- & Full-Think & 85.0 & 9.1k & 70.0 & 23.1k & 54.5 & 15.9k & 21.6 & 16.1k & 23.5 & 17.6k & 40.9 & 16.4k \\
-- & No-Think & 86.8 & 1.7k & 53.3 & 12.8k & 53.0 & 3.1k & 35.9 & 5.0k & 32.8 & 4.8k & 42.4 & 5.5k \\
\midrule
\multicolumn{14}{l}{\emph{Prompt-Tuning}} \\
\midrule
\PT{} & TF & 85.4 & 6.5k & 80.0 & 20.2k & 63.1 & 12.2k & 29.9 & 11.7k & 29.5 & 11.2k & 47.6 & 12.4k \\
\PT{} & SFT & 85.6 & 5.2k & 80.0 & 17.5k & 63.4 & 10.1k & 29.5 & 9.8k & 30.5 & 9.5k & 47.8 & 10.4k \\
\PT{} & DPO & 85.8 & 4.8k & 80.0 & 16.3k & 64.1 & 9.4k & 31.1 & 9.2k & 30.5 & 8.7k & 48.3 & 9.7k \\
\PT{} & GRPO & 86.2 & 4.1k & 80.0 & 14.8k & 63.0 & 8.2k & 30.1 & 7.8k & 30.2 & 7.5k & 47.9 & 8.5k \\
\midrule
\multicolumn{14}{l}{\emph{Routing}} \\
\midrule
\RT{} & TF & 85.8 & 6.2k & 73.3 & 22.4k & 52.0 & 11.1k & 29.9 & 15.1k & 29.2 & 16.8k & 44.1 & 14.3k \\
\RT{} & SFT & 85.4 & 4.5k & 73.3 & 15.8k & 53.5 & 7.6k & 29.9 & 10.2k & 30.4 & 11.3k & 44.5 & 9.9k \\
\RT{} & DPO & 85.6 & 3.8k & 73.3 & 12.4k & 54.0 & 6.2k & 30.5 & 8.5k & 30.6 & 9.1k & 44.8 & 8.0k \\
\RT{} & GRPO & 85.8 & 2.1k & 73.3 & 7.1k & 54.0 & 3.9k & 30.5 & 5.8k & 29.4 & 6.2k & 44.6 & 5.0k \\
\midrule
\multicolumn{14}{l}{\emph{Speculative}} \\
\midrule
\Spec{} & TF & 84.8 & 11.1k & 63.3 & 32.7k & 55.6 & 18.4k & 43.1 & 20.8k & 32.2 & 23.0k & 45.8 & 21.2k \\
\Spec{} & SFT & 85.8 & 6.9k & 63.3 & 28.1k & 53.0 & 13.5k & 36.5 & 15.3k & 34.8 & 16.1k & 44.7 & 16.0k \\
\Spec{} & DPO & 86.4 & 6.2k & 63.3 & 26.5k & 53.5 & 12.8k & 37.7 & 14.1k & 35.2 & 14.8k & 45.2 & 14.9k \\
\Spec{} & GRPO & 86.0 & 5.8k & 63.3 & 25.2k & 52.5 & 12.2k & 37.1 & 13.5k & 35.5 & 14.0k & 44.9 & 14.1k \\
\bottomrule
\end{tabular}
\caption{Full per-dataset training results (Qwen3.5-9B). Acc: Pass@1 (\%); Tok: average output tokens.}
\label{tab:training_full}
\vspace{-10pt}
\end{table*}

\section{Failure Case Analysis}
\label{app:failures}

We identify representative failure modes for each strategy:

\paragraph{PT failure: Over-compression on hard problems.} On AIME problems, PT-GRPO occasionally produces overly abbreviated reasoning (avg 4.1k tokens on MATH500 vs.\ 6.5k for PT-TF) that skips critical intermediate steps. While accuracy is maintained on average, per-problem inspection reveals that the model sometimes ``summarizes'' rather than fully reasons through multi-step proofs, occasionally leading to errors on problems requiring 4+ reasoning steps.

\paragraph{RT failure: Difficulty miscalibration on GPQA.} Our router achieves only 52.0\% accuracy on GPQA-Diamond, compared to 54.5\% for Full-Think. Inspection reveals that the router systematically \emph{overestimates} difficulty of science questions (routing 93\% to think-mode), yet the think-mode response still fails. The issue is not routing quality but the model's intrinsic inability on this task---routing cannot improve performance when neither mode produces correct answers.

\paragraph{Spec failure: Confident wrong answers.} On AIME, the speculative trigger's entropy-based mechanism fails when the model produces a \emph{confident but incorrect} initial answer. With Spec-Entropy, all 30 problems enter ``mixed'' mode (attempting retrigger), but the retrigger adds tokens without correcting the fundamental reasoning error. The trigger fires too late---after the model has already committed to a wrong approach.

\section{Implementation Details}
\label{app:impl_details}

This appendix provides comprehensive implementation details for all methods in HRBench, complementing the overview in \S\ref{sec:baselines}.

\subsection{Training Hyperparameters}
\label{app:hyperparams}

\begin{table}[h]
\centering
\small
\begin{tabular}{l|ccc}
\toprule
\textbf{Hyperparameter} & \textbf{SFT} & \textbf{DPO} & \textbf{GRPO} \\
\midrule
Learning rate & 2e-5 & 5e-7 & 5e-7 \\
Batch size & 4 & 4 & 4 \\
Epochs & 2 & 1 & 1 \\
Max length & 122880 & 122880 & 122880 \\
KL coefficient & -- & 0.1 & 0.01 \\
GRPO $n$ & -- & -- & 8 \\
Temperature & -- & -- & 1.0 \\
Framework & verl & verl & verl \\
\bottomrule
\end{tabular}
\caption{Training hyperparameters.}
\vspace{-10pt}
\label{tab:hyperparams}
\end{table}

\subsection{Model-Specific Parameter Mapping}
\label{app:api_mapping}

A key design principle of HRBench is that each strategy maps to model-native parameters rather than using a one-size-fits-all approach. 

\paragraph{Inference configuration.} All experiments use greedy decoding (temperature=0) except RASC (temperature=0.7 for diversity sampling). Maximum output tokens are set to 32,768 for all strategies. The judge stage in RT-TF uses max\_tokens=256 with no-think/low-effort mode to minimize overhead. The mode-decision heuristic post-generation classifies outputs as ``think'' if thinking\_tokens $> 100$, ``nothink'' if thinking\_tokens $< 10$, and ``brief\_think'' otherwise.

\subsection{Unified Training Data Construction}
\label{app:data_construction}

All training-based methods (SFT, DPO, GRPO) across all three strategies follow a unified rejection fine-tuning (RFT) pipeline. Given a training set from MathLightEval \cite{hendrycksmath2021}, our data construction proceeds as follows:

\paragraph{Step 1: Multi-mode rollout.} For each problem $q$, we generate $K$ responses under each available thinking mode $m \in \mathcal{M}$ (e.g., \texttt{think}, \texttt{no\_think}, or intermediate effort levels). Each rollout produces a response $r_{k,m}$ with associated token count $t_{k,m}$ and correctness label $c_{k,m} \in \{0, 1\}$.

\paragraph{Step 2: SFT sample selection.} We select the response that is both correct and token-minimal:
\begin{equation}
r^* = \arg\min_{r_{k,m}} t_{k,m} \quad \text{s.t.} \quad c_{k,m} = 1
\end{equation}
This yields training pairs $(q, r^*)$ that teach the model to produce efficient yet accurate responses.

\paragraph{Step 3: DPO pair construction.} The RFT process naturally produces preference pairs. The chosen response $r^+$ is the correct response with minimum token cost; rejected responses $r^-$ are drawn from (a) correct but longer responses, or (b) incorrect responses. Formally:
\begin{equation}
r^+ = r^*, \quad r^- \in \{r_{k,m} \mid t_{k,m} > t^* \text{ or } c_{k,m} = 0\}
\end{equation}

\paragraph{Step 4: GRPO reward design.} For GRPO, we perform $n=8$ rollouts per problem and compute a composite reward:
\begin{equation}
\begin{aligned}
R(r) = \alpha \cdot \mathbb{1}[\text{correct}(r)] + \\
\beta \cdot \mathbb{1}[\text{correct}(r)] \cdot \max\!\left(0, 1 - \frac{t_r}{t_{\text{ref}}}\right)
\end{aligned}
\end{equation}
where $t_{\text{ref}}$ is the Full-Think reference token count, $\alpha=1.0$ weights accuracy, and $\beta=0.5$ weights efficiency. Critically, the efficiency term is \emph{gated by correctness}: only correct responses receive efficiency bonuses, preventing degenerate compression. Per-group advantages are computed relative to the group mean reward across $n=8$ rollouts.

\paragraph{Prompt consistency.} A key design choice: training-based strategies (PT-SFT, PT-DPO, PT-GRPO, RT-SFT, etc.) use the \emph{identical} model-specific system prompt during both training data construction and inference.

\subsection{Prompt-Tuning Implementations}
\label{app:pt_impl}

\paragraph{PT-TF: Model-specific prompt design.} Each hybrid-reasoning model exposes a different set of thinking mode controls. We design model-specific prompts (full text in Appendix~\ref{app:prompts} \S B.1) that leverage the native mode interface:
\begin{itemize}[nosep,leftmargin=*]
\item \textbf{Qwen3.5 series}: The prompt guides the model to modulate its \texttt{<think>} block depth---empty/brief for simple problems, thorough for complex ones. 
\item \textbf{gpt-oss}: The prompt teaches mapping from difficulty assessment to three reasoning effort levels. 
\item \textbf{Seed-OSS}: The prompt teaches the model to self-allocate thinking budget and use the native \texttt{<seed:cot\_budget\_reflect>} tag for progress monitoring. 
\end{itemize}
In all cases, the model autonomously decides the reasoning depth---the prompt provides the decision framework but does not hard-code the choice. After generation, we classify output mode based on thinking token count: \texttt{think} ($>100$ tokens), \texttt{nothink} ($<10$ tokens), or \texttt{brief\_think} (intermediate).

\paragraph{PT-SFT.} Training data is constructed per Step 1--2 above using the model-specific PT system prompt: for each problem, we sample under both think and no-think modes, then select the correct response with minimum tokens. The model learns to internalize the prompt-guided mode selection.

\paragraph{PT-DPO.} Preference pairs are constructed per Step 3: the efficient correct response is chosen over verbose correct or incorrect alternatives. This teaches the model to prefer concise reasoning when prompt-guided.

\paragraph{PT-GRPO.} The model generates multiple responses under the PT prompt, receiving rewards per Step 4 that encourage both correctness and brevity.

\subsection{Routing Implementations}
\label{app:rt_impl}

\paragraph{RT-TF: LLM-as-router (two-stage).} The reasoning LLM itself serves as a router:
\begin{enumerate}[nosep,leftmargin=*]
\item \textbf{Stage 1 (Judge):} A lightweight call with model-specific judge prompt (Appendix~\ref{app:prompts} \S B.2). The judge runs in no-think/low-effort mode (\texttt{max\_tokens=256}) to minimize overhead. It outputs a JSON object specifying the routing decision:
    \begin{itemize}[nosep,leftmargin=*]
    \item Qwen3.5: \texttt{\{"mode": "1"|"2"|"3", "budget": N\}} (think / nothink / budget-think)
    \item gpt-oss: \texttt{\{"level": "high"|"medium"|"low"\}}
    \item Seed-OSS: \texttt{\{"mode": "1"|"2"|"3", "budget": N\}} (same format as Qwen)
    \end{itemize}
\item \textbf{Stage 2 (Solve):} The problem is dispatched according to the judge's decision.
\end{enumerate}
JSON parsing includes fallback: if the response is malformed, the conservative default (full think) is used. No additional parameters are trained---the LLM's existing capabilities drive the routing decision.

\paragraph{RT-SFT \& RT-DPO.} For both RT-SFT and RT-DPO, we collect routing labels as follows:
\begin{enumerate}[nosep,leftmargin=*]
\item For each problem $q$, run RFT under all available modes $m \in \mathcal{M}$.
\item Identify the mode $m^*$ that produces correct answers with minimum average token cost.
\item The routing label for $q$ is $m^*$.
\end{enumerate}
This produces router training samples $(q, m^*)$ for SFT, and preference pairs $(q, m^+, m^-)$ for DPO where $m^+ = m^*$ and $m^-$ is any alternative mode.

\paragraph{RT-GRPO.} During GRPO training, \emph{only the router is updated}---the backbone LLM is frozen. The router makes mode decisions, the LLM generates responses under the routed mode, and rewards are computed based on the final answer's correctness and token efficiency. This allows the router to learn optimal dispatching without modifying the LLM's reasoning capabilities.

\subsection{Speculative Implementations}
\label{app:spec_impl}

Both speculative variants follow a \textbf{two-pass architecture}:

\paragraph{Spec-Trigger: Keyword-based mode escalation.}
\begin{enumerate}[nosep,leftmargin=*]
\item \textbf{Pass 1:} Generate complete response in no-think mode.
\item \textbf{Decision:} Scan the response text for any match in the uncertainty keyword library (55 keywords across 6 categories; full list in \S B.3).
\item \textbf{Pass 2 (if triggered):} Discard Pass 1 output; re-generate with full think mode. Total token count = Pass 1 tokens + Pass 2 tokens.
\end{enumerate}
The keyword library is model-specific to account for different hedging patterns:
\begin{itemize}[nosep,leftmargin=*]
\item \textbf{Qwen3.5}: 55 keywords including \emph{wait}, \emph{actually}, \emph{let me reconsider}, \emph{I'm not sure}, \emph{hmm}, \emph{alternatively}, \emph{let me verify}, etc.
\item \textbf{gpt-oss}: Same core library; models tend to use \emph{hold on}, \emph{let me think again}.
\item \textbf{Seed-OSS}: Same core library; models tend to use \emph{let me re-examine}, \emph{actually no}.
\end{itemize}

\paragraph{Spec-Entropy: Token-level uncertainty trigger.}
\begin{enumerate}[nosep,leftmargin=*]
\item \textbf{Pass 1:} Generate complete response in no-think mode \textbf{with logprobs} (top-20 logprobs).
\item \textbf{Decision:} Compute normalized Shannon entropy for each output token:
\begin{equation}
H_t = \frac{-\sum_{v \in \text{top-}k} \hat{p}_t(v) \log \hat{p}_t(v)}{\log k}
\end{equation}
where $\hat{p}_t$ is the renormalized distribution over the top-$k=20$ tokens. Escalation fires if $\geq 3$ tokens or $>5\%$ of total output tokens exceed the model-specific threshold $\tau$.
\item \textbf{Pass 2 (if triggered):} Re-generate with full think mode. Total token count includes both passes.
\end{enumerate}

\paragraph{Spec-SFT/DPO.} Training data follows the same RFT pipeline: we generate responses under both the initial no-think pass and the full speculative (trigger/entropy $\to$ re-think) pipeline, then select correct responses with minimum total tokens as SFT targets. For DPO, the efficient correct response is chosen over alternatives that either triggered unnecessarily (wasting tokens) or failed to trigger when needed (producing wrong answers).

\paragraph{Spec-GRPO.} During training, the speculative mechanism runs end-to-end: the model begins in no-think mode, may trigger re-thinking, and produces a final answer. Multiple rollouts per problem receive rewards based on both answer correctness and total token cost (including any re-think overhead). The model learns when triggering is beneficial versus costly.

\subsection{External Method Reproduction}
\label{app:external_impl}

\paragraph{Prompt-Tuning methods.}
\begin{itemize}[nosep,leftmargin=*]
\item \textbf{S1} \citep{muennighoff2025s1}: Budget forcing via \texttt{thinking\_budget} API parameter. Budget levels: Low=1024, Medium=4096, High=16384 tokens. We report the High variant. For Qwen3.5, this maps directly to \texttt{thinking\_budget=16384}; for Seed-OSS, the same parameter is used.
\item \textbf{TALE} \citep{han2025tale}: Token-budget-aware explicit planning. The model first estimates a token budget (e.g., simple=100, medium=500, hard=1500), then reasons within that budget. We use the EP (self-Estimation Planning) variant with thinking enabled.
\item \textbf{Budget-Guidance} \citep{li2025steering}: Explicit token budget specified in system prompt. Budget levels: Low=128, Medium=512, High=2048. We evaluate the Medium variant. For Qwen3.5, we additionally set \texttt{thinking\_budget} as a soft constraint matching the prompt budget.
\item \textbf{SoT} \citep{aytes2025sketch}: Sketch-of-Thought with 3 cognitive paradigms (Chunked Symbolism / Conceptual Chaining / Expert Lexicons). We use domain-based paradigm selection (math$\to$Chunked Symbolism, science$\to$Conceptual Chaining, code$\to$Expert Lexicons). Runs in no-think mode since conciseness is prompt-driven.
\item \textbf{CoD} \citep{xu2025chain}: Chain-of-Draft with domain-specific compressed prompts instructing $\leq$5 words per reasoning step. Runs in no-think mode.
\item \textbf{DynaThink} \citep{pan2024dynathink}: Three-stage (fast generation $\to$ confidence probe $\to$ optional re-generation). Confidence threshold=0.7. 
\item \textbf{DEER} \citep{yang2025dynamic}: Dynamic early exit monitoring 9 transition patterns (\emph{Wait}, \emph{Alternatively}, \emph{Actually}, \emph{Let me reconsider}, \emph{On second thought}, \emph{Hmm}, \emph{No,}, \emph{But wait}, paragraph breaks). Confidence threshold=0.85, minimum thinking tokens=50. Uses logprobs (top-10) to compute per-token confidence at transition points.
\item \textbf{RASC} \citep{wan2025reasoning}: Reasoning-aware self-consistency. Configuration: max\_samples=8, min\_samples=3 (early stopping), consistency\_threshold=0.6, temperature=0.7. Scoring: 0.7$\times$consistency + 0.3$\times$brevity. Generates in think mode for quality.
\end{itemize}

\paragraph{Routing methods.}
\begin{itemize}[nosep,leftmargin=*]
\item \textbf{AdaptThink} \citep{zhang2025adaptthink}: GRPO-trained model that internally decides think/nothink. At inference, we simply enable thinking and let the trained model choose. Mode is inferred from output: $>50$ thinking tokens $\to$ think; $<5$ $\to$ nothink. The model's RL training (with $\delta$ parameter controlling thinking ratio) internalizes the routing decision.
\item \textbf{HDFlow} \citep{yao2024hdflow}: Rule-based difficulty routing using query complexity heuristics. We faithfully implement their classification rules and route to think/nothink accordingly.
\end{itemize}

\paragraph{Speculative methods.}
\begin{itemize}[nosep,leftmargin=*]
\item \textbf{MixReasoning} \citep{lu2025mixreasoning}: Entropy-based mode escalation during generation. We use their recommended entropy threshold and implement the same two-pass architecture as our Spec-Entropy.
\item \textbf{ADR} \citep{zhang2025adr}: SFT+GRPO-trained adaptive switching policy. We retrain following their pipeline on our training set for fair comparison.
\end{itemize}

\end{document}